\documentclass[conference]{IEEEtran}
\usepackage{times}

\usepackage[numbers]{natbib}
\usepackage{multicol}
\usepackage[bookmarks=true]{hyperref}

\usepackage{graphics} 
\usepackage{times} 
\usepackage{amsmath} 
\usepackage{amssymb}  
\usepackage{mathrsfs}
\usepackage{amsfonts}
\usepackage{kantlipsum}
\usepackage{subfigure}
\usepackage[ruled,linesnumbered]{algorithm2e}

\usepackage{hyperref}
\usepackage{graphicx}
\usepackage{float}
\usepackage{ctable}
\usepackage{cuted}
\usepackage{colortbl}
\definecolor{Gray}{gray}{0.9}
\definecolor{White}{gray}{1}
\usepackage{multirow}

\IEEEoverridecommandlockouts
\begin{document}
\title{\LARGE 
{\bf SAM-RL}: {\bf S}ensing-{\bf A}ware {\bf M}odel-Based {\bf R}einforcement {\bf L}earning via 
Differentiable Physics-Based Simulation and Rendering}

\author{Jun Lv$^{1}$, Yunhai Feng$^{2}$, Cheng Zhang$^{3}$, Shuang Zhao$^{3}$, Lin Shao$^{4*}$ and Cewu Lu$^{5*}$
\thanks{*Equal advising.}
\thanks{$^{1}$Jun Lv is with the Department of Electronic Engineering, Shanghai Jiao Tong University, China. \emph{[lyujune\_sjtu@sjtu.edu.cn]}}%
\thanks{$^{2}$Yunhai Feng is with the Department of Computer Science and Engineering, University of California San Diego, USA. \emph{[yuf020@ucsd.edu]}}%
\thanks{$^{3}$Cheng Zhang and Shuang Zhao are with the Department of Computer Science, University of California Irvine, USA. \emph{ [chengz20@uci.edu, shz@ics.uci.edu]}}%
\thanks{$^{4}$Lin Shao is with the Department of Computer Science, National University of Singapore, Singapore. \emph{[linshao@nus.edu.sg]}}%
\thanks{$^{5}$Cewu Lu is the corresponding author, the member of Qing Yuan Research Institute and MoE Key Lab of Artificial Intelligence, AI Institute,
Shanghai Jiao Tong University, China and Shanghai Qi Zhi Institute. \emph{[lucewu@sjtu.edu.cn]}}
}

\maketitle

\begin{abstract}
Model-based reinforcement learning (MBRL) is recognized with the potential to be significantly more sample efficient than model-free RL.
How an accurate \emph{model} can be developed automatically and efficiently from raw sensory inputs (such as images), especially for complex environments and tasks, is a challenging problem that hinders the broad application of MBRL in the real world.
In this work, we propose a sensing-aware model-based reinforcement learning system called \emph{SAM-RL}.
Leveraging the differentiable physics-based simulation and rendering, \emph{SAM-RL} automatically updates the model by comparing rendered images with real raw images and produces the policy efficiently. With the sensing-aware learning pipeline, \emph{SAM-RL} allows a robot to select an informative viewpoint to monitor the task process. We apply our framework to real world experiments for accomplishing three manipulation tasks: robotic assembly, tool manipulation, and deformable object manipulation.
We demonstrate the effectiveness of \emph{SAM-RL} via extensive experiments.
Videos are available on our project webpage at~\href{https://sites.google.com/view/rss-sam-rl}{https://sites.google.com/view/rss-sam-rl}.

\end{abstract}

\IEEEpeerreviewmaketitle

\section{Introduction}
Over the past decade, deep reinforcement learning~(RL) has resulted in impressive successes, including mastering Atari games~\cite{mnih2015human}, winning the games of Go~\cite{44806}, and solving Rubik's cube with a human-like robot hand~\cite{akkaya2019solving}.
However, deep RL algorithms adopt the paradigm of model-free RL and require vast amounts of training data, 
significantly limiting their practicality for real-world robotic tasks.
Model-based reinforcement learning~(MBRL) is recognized with the potential to be significantly more sample efficient than model-free RL~\cite{wang2019benchmarking}. 
\begin{figure}[t!]
 \centering
 \includegraphics[width=1.0\linewidth]{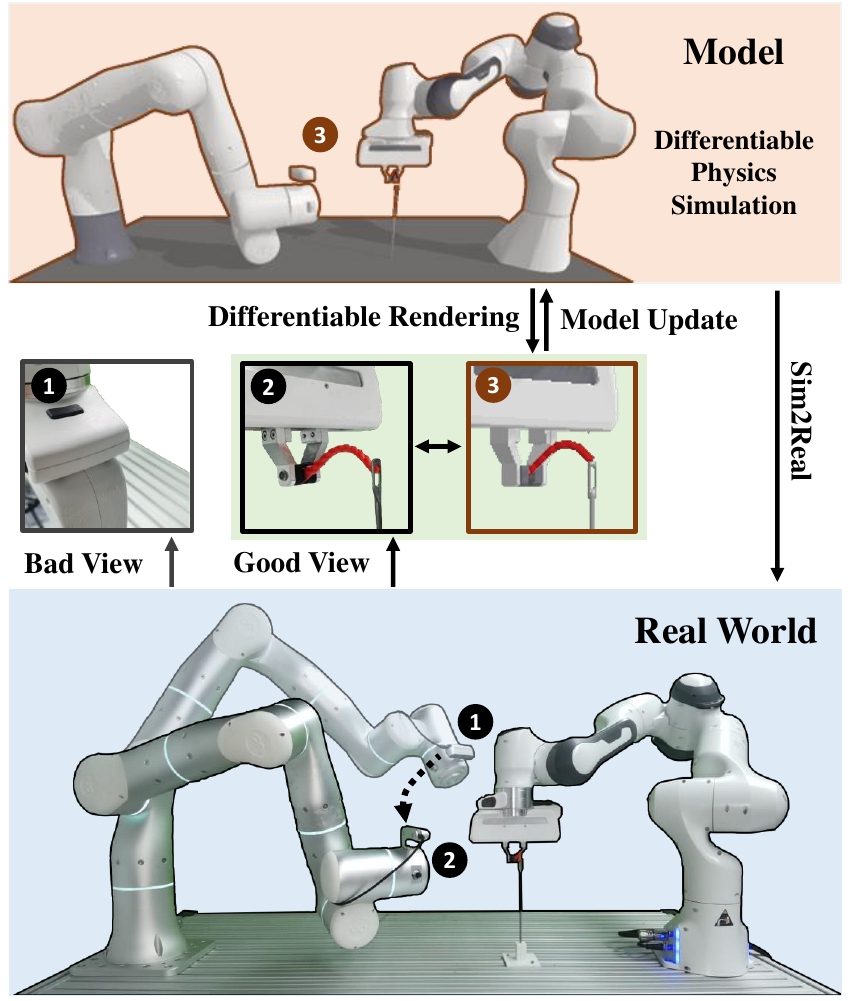}
  \caption{Our proposed \emph{SAM-RL} enables robots to autonomously select an informative camera viewpoint to better monitor the manipulation task~(for example, the \emph{Needle-Threading} task). We leverage the differentiable rendering to update the \emph{model} by comparing the raw observation between simulation and the real world, and differentiable physics simulation to produce policy efficiently.}
\label{fig:teaser}
\end{figure}

How to automatically and efficiently develop an accurate \emph{model} from raw sensory inputs, especially for complex environments and tasks, is a challenging problem that hinders the wide application of MBRL in the physical world. 


One line of works~\cite{yang2021representation,hafner2019dream,hafner2019learning,hafner2020dreamerv2} adopt representation learning approaches to learn the \emph{model} from raw input data. They aim to learn low-dimensional latent states and action representations from high-dimensional input data like images.
But the learned deep network might not satisfy the physical dynamics, and its quality may also significantly degenerate beyond the training data distribution when testing in the wild.
Recent developments in differentiable physics-based simulation~\cite{hu2019difftaichi,werling2021fast,howell_dojo_2022,du2021_diffpd,li2022diffcloth,qiao2021differentiable,qiao2020scalable,liang2019differentiable} and  rendering~\cite{liu2019soft,Li:2018:DMC,Zhang:2020:PSDR,Bangaru:2020:WAS} provide an alternative direction to model the environment~\cite{gradsim,ma2021risp}.
\citet{lv2022sagci} use differentiable physics-based simulation as the backbone of the \emph{model} and train robots to perform articulated object manipulation in the real world.
Their pipeline produces a file of Unified Robot Description Format~(URDF)~\cite{urdf} of the environment, which is loaded into the differentiable simulation from raw point clouds gathered by an RGB-D camera mounted on its wrist.
However, a sequence of camera poses is needed to scan the 3D environment every time step, and these camera poses are manually predefined, which is time-consuming and difficult to adapt to various tasks.
Object colors and 
geometric details are not included in the \emph{model}, limiting its representation capability~\cite{lv2022sagci}.

By integrating 
differentiable physics-based simulation and rendering, 
we propose a sensing-aware model-based reinforcement learning system called \emph{SAM-RL}. As shown in Fig.~\ref{fig:teaser}, we apply \emph{SAM-RL} on a robot system with two 7-DoF robotic arms (Flexiv Rizon~\cite{Flexiv} and Franka Emika Panda), where the former mounts an RGB-D camera, and the latter handles manipulation tasks. Our framework is sensing-aware, which allows the robot to automatically select an informative camera view to effectively monitor the manipulation process, providing 
the following benefits. First, the system no longer 
 requires obtaining a sequence of camera poses at each step, which is extremely time-consuming.
Second, compared with using a fixed view, \emph{SAM-RL} leverages varying camera views with potentially fewer occlusions and offers better estimations of environment states and object status (especially for deformable bodies). The improved quality in object status estimation contributes more effective robotic actions to complete various tasks.
Third, by comparing rendered and measured (i.e., real-world) images, discrepancies between the simulation and the reality are better revealed and then reduced automatically using gradient-based optimization and differentiable rendering.

In practice, we train the robot to learn three challenging manipulation skills: \emph{Peg-Insertion}, \emph{Spatula-Flipping}, and \emph{Needle-Threading}.
Our experiments indicate that \emph{SAM-RL} can significantly reduce training time and improve success rate by large margins compared to common model-free and model-based deep reinforcement learning algorithms.

Our primary contributions include:
\begin{itemize}
    \item proposing an active-sensing framework named \emph{SAM-RL} that enables robots to select informative views for various manipulation tasks;
    \item introducing a model-based reinforcement learning algorithm to produce efficient policies;
    \item conducting extensive quantitative and qualitative evaluations to demonstrate the effectiveness of our approach;
    \item applying our framework to robotic assembly, tool manipulation, and deformable object manipulation tasks both in simulation and real world experiments.
\end{itemize}


\section{Related Work}
We review related literature on key components in our approach, including model-based reinforcement learning, next best view, integration of differentiable physics-based simulation and rendering, and robotic manipulation. We describe how we are different from previous work.

\subsection{Model-based Reinforcement Learning}
MBRL is considered to be potentially more sample efficient than model-free RL~\cite{wang2019benchmarking}. However, automatically
and efficiently developing an accurate \emph{model} from raw sensory
data is a challenging problem, which retards MBRL from being widely applied in the real world. For a broader review of the field on MBRL, we refer to~\cite{luo2022survey}. 
One line of works~\cite{yang2021representation,hafner2019dream,hafner2019learning,hafner2020dreamerv2} use representation learning methods to learn low-dimensional latent state and action representations from high-dimensional input data. But the learned models might not satisfy the physical dynamics, and the quality may also significantly drop beyond the training data distribution. Recently, ~\citet{lv2022sagci} leveraged the differentiable physics simulation and developed a system to produce a URDF file to model the surrounding environment based on an RGB-D camera. However, the RGB-D camera poses used in~\cite{lv2022sagci} are predefined and can not adjust to different tasks. Our approach allows the robot to select the most informative camera view to monitor the manipulation process and update the environment \emph{model} automatically.

\subsection{Next Best View in Active Sensing}
Next Best View~(NBV) has been one of the core problems in active sensing. It studies the problem of how to obtain a series of sensor poses to increase the information gain. The information gain is explicitly defined to reflect the improved perception for 3D reconstruction~\cite{collander2021learning,5980429,peralta2020next,kriegel2015efficient}, object recognition~\cite{wu20143d,9811769,5152350,6840371}, 3D model completion~\cite{5979947}, and 3D exploration~\cite{SURMANN2003181,bircher2016receding}. Unlike perception-related tasks, we explore the NBV over a wide range of robotic manipulation tasks. Information gain in the robotic manipulation tasks is difficult to define explicitly and is implicitly related to task performance. In our system, the environment changes accordingly after the robot's interaction. We integrate the information gain into the $Q$ function to reflect the informative viewpoint for manipulation. 

\subsection{Integration of Differentiable Physics-Based Simulation and Rendering}
Recently, great progresses have been made in the field of differentiable physics-based simulation and rendering. For a broader review, please refer to~\cite{hu2019difftaichi,werling2021fast,howell_dojo_2022,du2021_diffpd,li2022diffcloth,qiao2021differentiable,qiao2020scalable,liang2019differentiable} and~\cite{kato2020differentiable,zhao2020physics}.
With the development of these techniques, \citet{gradsim} first proposed a pipeline to leverage differentiable simulation and rendering for system identification and visuomotor control. \citet{ma2021risp} introduced a rendering-invariant state predictor network that maps images into states that are agnostic to rendering parameters. By comparing the state predictions obtained using rendered and ground-truth images, the pipeline can backpropagate the gradient to update system parameters and actions. \citet{sundaresan2022diffcloud} proposed a real-to-sim parameter estimation approach from point clouds for deformable objects.
 Different from these works, we use the differentiable simulation and rendering to find the next best view for various manipulation tasks and update the object status in the \emph{model} by comparing  rendered and captured images.

\begin{figure*}[t!]
  \begin{center}
   \includegraphics[width=1.0\linewidth]{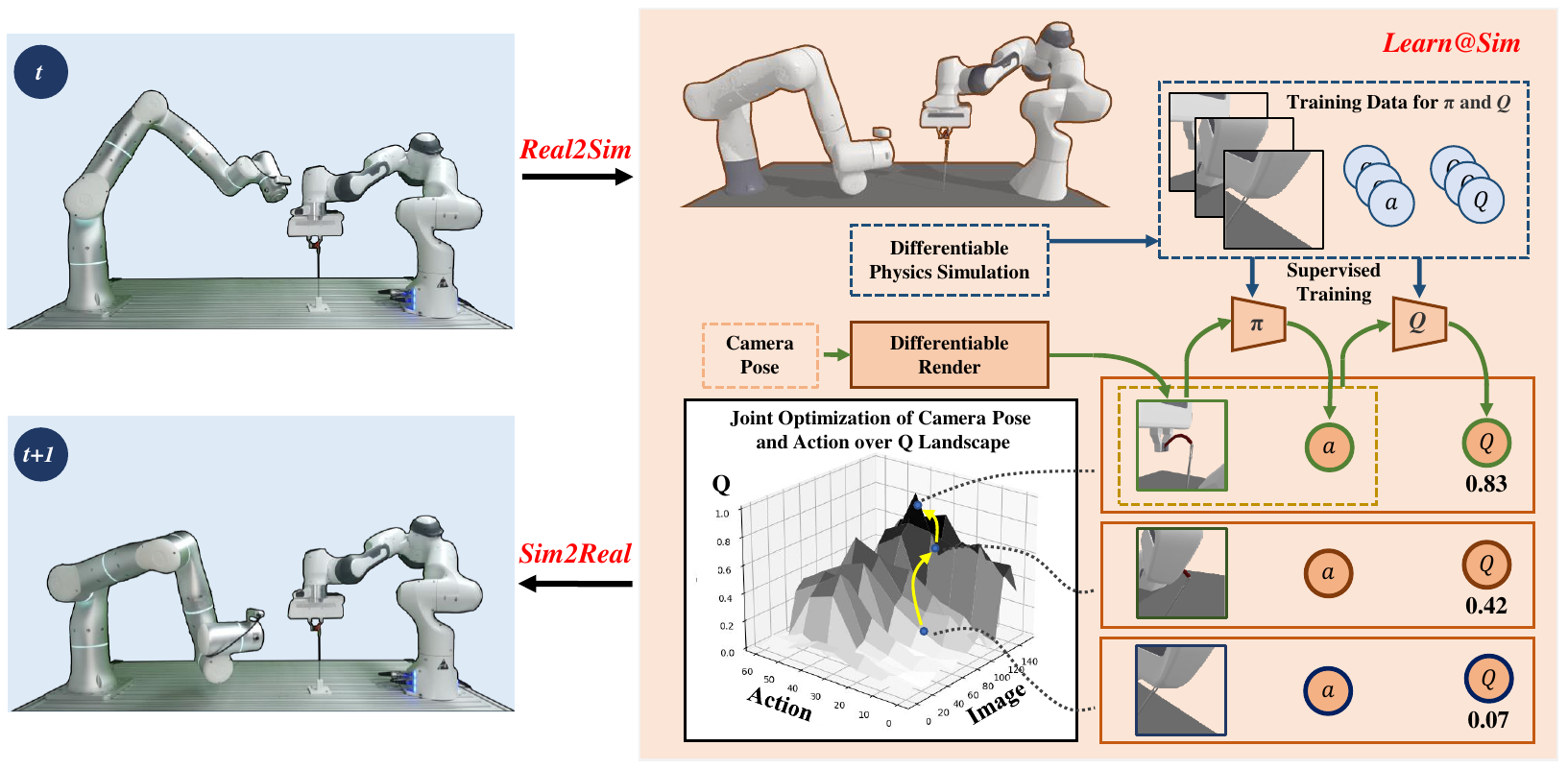}
  \end{center}
      \caption{The overall approach of \emph{SAM-RL} includes Real2Sim, Learn@Sim, and Sim2Real stages. \emph{SAM-RL} automatically develop and update the \emph{model} during Real2Sim stage. During the Learn@Sim stage, it learns the sensing-aware $Q$ function and actor~$\pi^{sim}$ in the \emph{model}. The differentiable physics simulation generates training data (rendered image, action, and associated return) to learn the $Q$ function and actor function, which allows the robot to select an informative view. In the Sim2Real stage, \emph{SAM-RL} learns a residual policy to reduce the sim-to-real gap.}
\label{fig:overview}
\end{figure*}
 
\subsection{Manipulation}
Our framework can be adopted to improve the performance of a range of manipulation tasks. We review the related work in these domains. \emph{1) Peg-insertion.} ~Peg insertion is a classic robotic assembly task with rich literature~\cite{luo2019reinforcement,lee2019making,shao2020learning}. For a broad review of peg-insertion, we refer to~\cite{xu2019compare}.  
\emph{2) Spatula-Flipping.} ~\citet{6943031} used the tactile sensor to train the robot learning to perform a scraping task with a spatula. ~\citet{7358120} studied the dynamic object manipulation by a spatula. They clarified the conditions for achieving dynamic movements and presented a unified algorithm for generating a variety of movements. 
\emph{3) Needle-Threading.}
The needle threading task requires the robots to adapt action according to the thread deformation.  \citet{silverio2021laser} relied on a high-resolution laser scanner to perceive the thread and needle. \citet{7353947} used a high-speed camera to monitor the process and provide high-speed visual feedback.~\citet{9354902} proposed a deep imitation learning algorithm for the needle threading task. Unlike approaches above, we develop a sensing-aware model-based reinforcement learning approach to learn these skills.

\section{Technical Approach}

Given a manipulation task denoted as $\mathcal{T}$, our pipeline takes as input images gathered from an RGB-D camera and outputs a policy to select the camera pose $\mathcal{P}^{c}$ 
followed by producing an action $a$. An overview of our proposed method is shown in Fig.~\ref{fig:overview}. 
In what follows, we first briefly introduce the \emph{model} $\mathcal{M}$ 
that integrates differentiable physics-based simulation and rendering in Sec.~\ref{sec:diffX}.
Then, we describe developing and updating the \emph{model} in $\mathcal{M}$~(Real2Sim) in Sec.~\ref{sec:real_sim}, training robots to learn the perception and action with the \emph{model}~(Learn@Sim) in Sec.~\ref{sec:learn}, and applying the learned \emph{model} to the real world~(Sim2Real) in Sec.~\ref{sec:sim_real}. 


\subsection{Model with Differentiable Simulation and Rendering}\label{sec:diffX}
In this work, we combine the differentiable physics-based simulation and rendering.
The resulting differentiable system plays the backbone of the \emph{model}, which we denote as $\mathcal{M}$, for the model-based reinforcement learning.
The \emph{model} can load robots, cameras, and objects denoted as $\{\mathcal{O}^{sim}_j\}$ along with their visual/geometric (e.g., shape, pose, and texture) and physical attributes (e.g., mass and inertial).
We denote the attributes of all objects loaded in the simulation as one type of \emph{model}'s parameters $\psi_{\mathcal{M}}$ as follows:
\begin{equation}
    \label{eqn:updateM}
    \psi_\mathcal{M} = \sum_{j} \mathcal{O}^{sim}_j.
\end{equation}

The \emph{model} can render an image $\mathcal{I}^{sim}$ under the camera pose $\mathcal{P}^{c}$ and \emph{model} parameter $\psi_{\mathcal{M}}$ through:
\begin{equation}
    \mathcal{I}^{sim} =\mathcal{M}(\psi_\mathcal{M},\mathcal{P}^{c}; \textit{render})
\end{equation}
We can get the gradients $\partial{\mathcal{I}^{sim}}/\partial{\mathcal{P}^{c}}$ and $\partial{\mathcal{I}^{sim}}/\partial{\mathcal{O}^{sim}_j}$ using differentiable rendering~\cite{Li:2018:DMC}.
Note that $\partial{\mathcal{I}^{sim}}/\partial{\mathcal{O}^{sim}_j}$ contains only the gradient 
with respect to object visual and geometric attributes.

Additionally, given the state $s^{sim}_t$ including the object attributes $\psi_{\mathcal{M}}$ and the robots' status, when an action denoted as $a^{sim}_t$ is executed (for example, an external force is exerted on the object), the \emph{model} 
simulates the next state via
\begin{equation}
    s^{sim}_{t+1}=\mathcal{M}(s^{sim}_t,a^{sim}_t; \textit{forward})
\end{equation} 
in a differentiable fashion~\cite{werling2021fast}, providing the gradients $\partial{s}^{sim}_{t+1}/\partial{a}^{sim}_t$ and $\partial{s}^{sim}_{t+1}/\partial{s}^{sim}_t$.

\subsection{Real2Sim: Developing Model from the Real World}\label{sec:real_sim}
\subsubsection{Build the Initial Model}\label{sec:init}
For model-based reinforcement learning, the first step is to build an initial \emph{model} of the environment. The initial \emph{model} does not need to be accurate and can be created using current 3D object reconstruction methods with RGB-D cameras like BundleFusion~\cite{dai2017bundlefusion} and KinectFuction~\cite{izadi2011kinectfusion}. In our setting, as shown in Fig.~\ref{fig:overview}, a calibrated RGB-D camera is mounted on the robot's wrist. Therefore the robot system takes a set of images with corresponding accurate camera poses. The initial \emph{model} can also be built directly from a CAD model~\cite{thomas2018learning} or following the pipeline described in SAGCI~\cite{lv2022sagci} to produce the URDF. After the initialization, the \emph{model} $\mathcal{M}$ contains the robots, objects and an RGB-D camera.

\subsubsection{Update the Model with Differentiable Siumulation and Rendering}\label{sec:updateM}
After having an initial \emph{model}, we then describe how to update the \emph{model} by directly comparing the raw visual observation in simulation and the real world, leveraging the differentiable simulation~\cite{werling2021fast} and rendering~\cite{Li:2018:DMC}. In this work, we only care about one object and assume we can get accurate object segmentation of real world images. With common techniques such as Mask R-CNN~\cite{8237584}, it’s feasible to get a fine object-level instance segmentation. 

At the beginning, we update the camera and robot pose in simulation to match the corresponding camera and robot pose in the real world. Then we get the rendered RGB-D image from $\mathcal{M}$ and corresponding real RGB-D image with associated segmentation. Based on the camera parameters, we transform the depth image to point cloud and segment the point cloud. We denote the RGB image, associated object segmentation, and segmented point cloud ($\mathcal{I}^{sim,rgb}$, $\mathcal{G}^{sim}$, $\mathcal{X}^{sim}$) in the simulation and ($\mathcal{I}^{real,rgb}$, $\mathcal{G}^{real}$, $\mathcal{X}^{real}$) in the real world. For simplicity, we use $\mathcal{I}^{sim}$ to represent the ($\mathcal{I}^{sim,rgb}$, $\mathcal{G}^{sim}$, $\mathcal{X}^{sim}$) and $\mathcal{I}^{real}$ is defined accordingly.


We define the loss functions as follows.
\begin{align}
\mathcal{L}_1 &= \|{\mathcal{G}}^{real} \otimes \mathcal{I}^{real,rgb}- {\mathcal{G}}^{sim}   \otimes \mathcal{I}^{sim,rgb}\|_1\\
\mathcal{L}_2 &= \textit{EMD}(\mathcal{X}^{sim},\mathcal{X}^{real})\\
\mathcal{L} &= \lambda_1 \mathcal{L}_1 + \lambda_2 \mathcal{L}_2 \label{eqn:updateloss}
\end{align}
where $\otimes$ represents the pixel-wise product operator and \textit{EMD} represents the Earth Mover Distance~\cite{fan2017point} to measure the distance between two 3D point clouds. Note that $(\mathcal{I}^{sim,rgb}, \mathcal{G}^{sim}, \mathcal{X}^{sim})=\mathcal{M}(\psi_\mathcal{M}, \mathcal{P}^{c};\textit{render})$, as explained in Sec.~\ref{sec:diffX}, is differentiable. 
With the gradient $\partial \mathcal{L}/\partial \psi_\mathcal{M}$, we can update the \emph{model} $\mathcal{M}$'s parameters $\psi_\mathcal{M}$ including object mesh vertices, colors, and poses so that the loss $\mathcal{L}$ is reduced.
\begin{equation}
\psi_\mathcal{M} \leftarrow \psi_\mathcal{M} - \lambda_\mathcal{M} \frac{\partial \mathcal{L}}{\partial \psi_\mathcal{M}} \label{equ:modelupdate}
\end{equation}

During the manipulation, at each time step $t$, the \emph{model} parameters like the poses would be changed by the robot action $a_t$. We denote the object poses as  $\mathcal{P}^\mathcal{O}_t$. We will first update the object pose via the forward simulation by
\begin{equation}
\mathcal{P}^\mathcal{O}_{t+1} \leftarrow \mathcal{M}(\mathcal{P}^\mathcal{O}_t, a_t, forward)
\label{eqn:simulate}
\end{equation}
However, the sim-to-real gap may make the \emph{model} pose in simulation inaccurate. So we still need to use the method described in~\ref{equ:modelupdate} to further update the \emph{model} pose.

Next, we describe how to update objects' physical attributes. Starting from poses $\mathcal{P}^\mathcal{O}_0$ at timestep $t=0$, with a sequence of actions $\{a_t\}_{t=0}^T$ applied, the simulation calculates the object poses at timestep $t =T$ denoted as $\mathcal{P}^\mathcal{O}_{T}$ based on objects' physical attributes by simulating Eqn.~\ref{eqn:simulate} for $T$ steps.
\begin{equation}
\mathcal{P}^\mathcal{O}_{T} \leftarrow \mathcal{M}(\mathcal{P}^\mathcal{O}_0, \{a_t\}_{t=0}^T,forward)
\end{equation}
The object poses from the real-world at timestep $t=T$ are gathered with differentiable rendering denoted as $\Tilde{\mathcal{P}}^\mathcal{O}_{T}$. We then calculate the distance between the two poses. 
\begin{equation}
    \mathcal{L}_p=\|\Tilde{\mathcal{P}}^\mathcal{O}_{T} -
 \mathcal{P}^\mathcal{O}_{T}\|
\end{equation} 
Then object physical attributes $\psi_\mathcal{M}^{phy}$ (mass and inertial) are updated to minimize the loss $\mathcal{L}_p$.
 \begin{equation}
\psi_\mathcal{M}^{phy} \leftarrow \psi_\mathcal{M}^{phy} - \lambda_\mathcal{M} \frac{\partial \mathcal{L}_p}{\partial \psi_\mathcal{M}^{phy}}
\end{equation}

Through this, we can keep reducing the discrepancy between the simulation and the real world, making the \emph{model} more and more accurate. Up to now, we have introduced how to build and update the \emph{model}.

\subsection{Sim2Real: Learning Residual Policy in the Real World}\label{sec:sim_real}
We delay the discussion of how to learn the policy to complete the task $\mathcal{T}$ with the \emph{model} in the simulation to the next subsection~\ref{sec:learn}. Here we assume that we have the policy $\pi^{sim}$ in simulation which takes rendered images $\mathcal{I}^{sim}$ as inputs and outputs actions denoted as
\begin{equation}
    a^{sim}=\pi^{sim}(\mathcal{I}^{sim})
\end{equation} We describe how to apply the learned policy from simulation to the real world by learning a residual policy $\pi^{res}$ to reduce the sim-to-real gap in this subsection.

We set the same camera pose denoted as $\mathcal{P}^{c}$ both in the simulation and the real world and get images denoted as $\mathcal{I}^{sim,0}$ and $\mathcal{I}^{real}$. We first update the \emph{model}'s parameter $\psi_\mathcal{M}$ by minimizing the loss with Eqn.~\ref{eqn:updateloss} as described in Sec.~\ref{sec:updateM} and then get the new images $\mathcal{I}^{sim}$ with the updated $\psi_\mathcal{M}$.

With the $\mathcal{I}^{sim}$ and $\mathcal{I}^{real}$, we get an action from simulation $a^{sim}$. Instead of directly applying the action $a^{sim}$ in the real world, we train a residual policy that takes in the real image $\mathcal{I}^{real}$ and the $a^{sim}$, outputs the residual action
\begin{equation}
    \delta a^{real} = \pi^{res}(\mathcal{I}^{real},a^{sim})
\end{equation} Then actor model in the real world gets the action as follows.
\begin{equation}
a^{real}=\pi^{real}(\mathcal{I}^{real},a^{sim})=a^{sim}+\delta a^{real}
\end{equation}

We follow the training process of residual policy described in~\cite{silver2018residual}. The residual policy should never make a good initial policy worse. We therefore initialize the residual policy so
that
\begin{equation}
    \pi^{res}(\mathcal{I}^{real}, a^{sim}) = 0
\end{equation}
before training. We do this by initializing the last layer of the network to be zero. Once the real world action $a^{real}$ is executed, we receive the next image $\mathcal{I}^{real'}$ and the reward
\begin{equation}
    r^{real} = 
    \begin{cases}
    1 & succeed\\
    0 & otherwise
    \end{cases}\label{eqn:reward}
\end{equation}  
The task will be considered \textit{done} when it succeeds or exceeds the max action step number. To train the residual policy $\pi^{res}$, we store the $(\mathcal{I}^{real},a^{real},a^{sim},r^{real},\textit{done},\mathcal{I}^{real'})$ to the reply buffer to update the $\pi^{res}$ following a common deep reinforcement learning procedure \emph{TD3}~\cite{fujimoto2018addressing}, an actor-critic deep RL method.

\subsection{Learn@Sim: Learning Sensing-aware Action in Simulation}~\label{sec:learn}

With the \emph{model}, we discuss how to learn the informative camera pose $\mathcal{P}^{c}$ associated with the rendered image $\mathcal{I}^{sim}$ and action $a^{sim}$ to complete the given task $\mathcal{T}$ in the simulation. 
\subsubsection{Sensing-aware Q-function} We adopt the $Q$ function $Q^{sim}(\mathcal{I}^{sim},a^{sim})$ to reflect the informative viewpoint.  We can calculate the gradient of $Q^{sim}$ with respect to the camera pose $\mathcal{P}^{c}$ as follows.
\begin{equation}
\frac{\partial{Q^{sim}}(\mathcal{I}^{sim},a^{sim})}{\partial \mathcal{P}^{c}}= \frac{\partial{Q^{sim}}(\mathcal{I}^{sim},a^{sim})}{\partial \mathcal{I}^{sim}} \frac{\partial{I^{sim}}}{\partial \mathcal{P}^{c}}  
\end{equation}
Here the first term $\partial{Q^{sim}}/\partial \mathcal{I}^{sim}$ is available through the backward propagation of the neural network. The second term $\partial{I^{sim}}/\partial \mathcal{P}^{c}$ can be obtained from the differentiable renderer~\cite{Li:2018:DMC}. The gradient $\partial{Q^{sim}}/\partial \mathcal{P}^{c}$ provides information on how to find a more informative viewpoint, which makes the pipeline sensing-aware. Under the more informative viewpoint, the actor has a better sense of the state to generate an action associated with a higher $Q$ value. We verify our hypothesis by visualizing the learned $Q$ function in Sec~\ref{sec:visQ}.


\subsubsection{Learning Actor in Simulation to Reflect the Sensing Quality in the Real World}\label{sec:actor} 
In this part, we discuss how we learn an actor $\pi^{sim}$ in simulation, which takes rendered image $\mathcal{I}^{sim}$ and outputs the action $a^{sim}$. 
There are multiple ways to learn the $\pi^{sim}$ in the simulation via reinforcement learning or imitation learning. In this work, we choose the imitation learning approach 
to learn the $\pi^{sim}$, which we find effective and efficient. With the \emph{model} $\mathcal{M}
$ built with the differentiable physics simulation~\cite{werling2021fast}, our pipeline generates trajectories completing the tasks $\{(s_t^{sim}, a_t^{sim} | \mathcal{T})\}_{t=1}^{T}$ to train the actor in simulation $\pi^{sim}$. To effectively generate the trajectories inside the differentiable physics simulation~\cite{werling2021fast}, we follow the method mentioned in~\cite{lv2022sagci}. Given task $\mathcal{T}$, we optimize the trajectory via the following equations.
\begin{align}\label{eqn:loss_abs}
 \mathcal{L}(\mathcal{T})&=\sum_{t=0}^{T-1}l(s^{sim}_t,a^{sim}_t;\mathcal{T})\\
 \label{eqn:loss_fw}
 \text{s.t.\hspace{2mm}} & s^{sim}_{t+1} = \mathcal{M}(s^{sim}_t,a^{sim}_t,forward)\\
 \label{eqn:loss_init}
   & s^{sim}_0 = s^{\text{init}}
\end{align}
Here we explain the loss function in Eqn.~\ref{eqn:loss_abs}.
\begin{align}
\label{eqn:loss_loss}
&l(s^{sim}_t,a^{sim}_t;\mathcal{T}) \\
&= \begin{cases}
\label{eqn:loss_sub}
\|a^{sim}_{t}\|^2 & t < T_1-1\\
\|s^{sim}_{T_1-1}-\mathcal{G}(\mathcal{T})_1\|^2 & t = T_1-1\\
\vdots\\
\|a^{sim}_{t}\|^2 & T_{n-1}-1 < t < T_n-1\\
\|s^{sim}_{T_n-1}-\mathcal{G}(\mathcal{T})_n\|^2 & t = T_n-1
\end{cases}
\end{align}
$\mathcal{G}(\mathcal{T})$ is the goal of the task. to make it easier to generate the trajectories, we define
$n$ sub-goals for each task. In practice, instead of directly optimizing the whole trajectory, the pipeline will optimize the trajectory to achieve each sub-goal in sequence. Once a goal is achieved, the pipeline will move on to the next goal.

After gathering these trajectories $\{(s_t^{sim}, a_t^{sim} | \mathcal{T})\}_{t=1}^{T}$ in the simulation, we rendered multiple images $\{\mathcal{I}^{sim}_t(\mathcal{P}^{c,i})\}$ for each state $s^{sim}_t$ under different camera poses $\{\mathcal{P}^{c,i}\}$. We train an actor $\pi^{sim}(\mathcal{I}_t(s_t^{sim},\mathcal{P}^{c,i}))$ to predict the action $\hat{a}^{sim}_t$ to imitate the corresponding action $a^{sim}_t$. In the process, the prediction error is defined as follows
\begin{equation}
\mathcal{L}^{\pi^{sim}}=\|\hat{a}_t^{sim}-a_t^{sim}\|_2
\end{equation}

During the learning process, the prediction error also depends on the sensing quality. Our hypothesis is the actor might not learn the effective ground-truth action if there is insufficient state information in the rendered image. For example, if the camera is always looking into the sky and the object does not appear in the rendered image, it will be difficult for the actor to learn the correct action. We visualize the prediction error in Sec.\ref{sec:visQ}. Note that the learning process also generates success and failure trajectories, which are used to learn the $Q$ function. 

\subsubsection{Learning Sensing-aware $Q$ Function} 
Here we describe how to learn the sensing-aware $Q$ function denoted as $Q^{sim}(\mathcal{I}^{sim}, a^{sim})$. There are also multiple ways to learn the $Q$ function. In this work, we formulate it as a supervised learning problem.  We roll out the trajectories in the simulation to generate the training data. Given a trajectory $\{(\mathcal{I}^{sim}_t(\mathcal{P}^{c,i}_t),\hat{a}^{sim}_t)\}_{t=1}^T$, we calculate the return  $\mathcal{R}_t=\sum_{i=t}^{T}\gamma^{T-i}r
_i$ associated with each image and action pair, where $r_t$ is the reward with same definition as Eqn.~\ref{eqn:reward} and $\gamma$ is the discount factor. We then train a deep network that takes $\mathcal{I}^{sim}_t$ and $\hat{a}^{sim}_t$ as inputs and outputs the $Q$ value by minimizing the following loss.
\begin{equation}
\mathcal{L}^Q = \| Q^{sim}(\mathcal{I}^{sim}_t,\hat{a}_t^{sim})-\mathcal{R}_t\|_2 
\end{equation}

\subsubsection{Selecting Perception and Action Leveraging the Actor and Q-function}~\label{sec:select} Starting from an initial viewpoint $\mathcal{P}^{c}$, the simulation rendered an image $\mathcal{I}^{sim}$. We feed the image into the actor model to get an action $a^{sim}=\pi^{sim}(\mathcal{I}^{sim})$, and then send the image and action to the $Q$ function to get the value $Q^{sim}(\mathcal{I}^{sim},a^{sim})$. We get the gradient with respect to the $\mathcal{P}^{c}$ and update the camera pose as follows.
\begin{equation}
 \mathcal{P}^{c'} = \mathcal{P}^{c} + \lambda \frac{\partial Q^{sim}(I^{sim},a^{sim})}{\partial \mathcal{P}^{c}}
\end{equation}\label{eqn:campose}
With the new camera pose $\mathcal{P}^{c'}$, we gather the new rendered image  $\mathcal{I}^{sim'}$, action $a^{sim'} = \pi^{sim}(\mathcal{I}^{sim'})$, and the associated $Q^{sim}(\mathcal{I}^{sim'},a^{sim'})$. We accept the new camera pose $\mathcal{P}^{c'}$ and new action $a^{sim'}$ if $Q(\mathcal{I}^{sim'},a^{sim'}) \geq Q(\mathcal{I}^{sim},a^{sim})$.  

\subsection{Overall Learning Process}

\begin{algorithm}[t]\label{alg:all}
\footnotesize
\caption{Overall algorithm}
\KwIn{the model $\mathcal{M}$ with its model parameters~$\psi_\mathcal{M}$, camera pose~$\mathcal{P}$, take the real image using real camera~\textit{Cam}($\mathcal{P}$)}
\For{$t$ in each iteration}{
   $\mathcal{I}^{sim} \leftarrow \mathcal{M}(\psi_\mathcal{M},\mathcal{P};render)$, $\mathcal{I}^{real} \leftarrow \textit{Cam}(\mathcal{P})$\;
   Update $\psi_\mathcal{M}$ by reducing $\|\mathcal{I}^{sim} - \mathcal{I}^{real}\|$\;
    $\mathcal{I}^{sim} \leftarrow \mathcal{M}(\psi_\mathcal{M},\mathcal{P};render)$, $a^{sim} = \pi^{sim}(\mathcal{I}^{sim})$\;
   \While{True}{
      $\mathcal{P}' = \mathcal{P} + \lambda_\mathcal{P} \frac{\partial Q(I^{sim},a^{sim})}{\partial\mathcal{P}}$\;
      $\mathcal{I}^{sim'} \leftarrow \mathcal{M}(\psi_\mathcal{M},\mathcal{P}';render)$, $\mathcal{I}^{real'} \leftarrow \textit{Cam}(\mathcal{P}')$\; 
      Update $\psi_\mathcal{M}$ by reducing $\|\mathcal{I}^{sim'} - \mathcal{I}^{real'}\|$\;
      $\mathcal{I}^{sim'} \leftarrow \mathcal{M}(\psi_\mathcal{M},\mathcal{P}';render)$,
     $a^{sim'} = \pi^{sim}(\mathcal{I}^{sim'})$\;
    \If{$Q^{sim}(\mathcal{I}^{sim},a^{sim}) \leq Q^{sim}(\mathcal{I}^{sim'},a^{sim'})$}
    {
     $\mathcal{I}^{real} \leftarrow \mathcal{I}^{real'}$, $a^{sim} \leftarrow a^{sim'}$\;
     break\;
   }
   $\mathcal{I}^{sim} \leftarrow \mathcal{I}^{sim'}, \mathcal{P} \leftarrow \mathcal{P}'$, $a^{sim} \leftarrow a^{sim'}$\;
   }
   \textbf{Add the updated Model with $\psi_\mathcal{M}$ to data buffer to generate trajectories}\;
   \textbf{Update the actor and Q function}\;
   $a^{real}=\pi^{real}(\mathcal{I}^{real},a^{sim})$\;
   Execute the action $a^{real}$ both in the real world and in simulation\;
   \textbf{Add the transition data into the replay buffer to sim2real}\;
   \textbf{Update the residual policy}
}
\end{algorithm}\label{alg:allt}
We summarize the overall test stage pipeline in the following Alg.~\ref{alg:all}. Given an initial \emph{model}, we leverage the differentiable rendering and simulation to update the \emph{model} parameter $\psi_{\mathcal{M}}$ and select the camera pose (Line 1-15). 

Our algorithm add the update \emph{model} $\mathcal{M}$ into our data buffer. Then our framework starts to generate the trajectories to accomplish the according task starting from the state described by the \emph{model} $\mathcal{M}$ associated with the parameter $\mathcal{M}$. (Line 16) 

We calculate the return value and expert actions to train the actor network and Q network in simulation. (Line 17)

Then we train the residual policy to deal with the sim2real gap. (Line 18-21)

\section{Experiments}
In this section, we introduce our experimental setup and conduct quantitative and qualitative evaluations to demonstrate the effectiveness of our approach. Our experiments focus on evaluating the following questions
\begin{itemize}
    \item Can our proposed selecting perception and action procedure described at~\ref{sec:select} improve the performance of various manipulation tasks?
    \item How does our \emph{SAM-RL} approach compare with model-free and model-based deep reinforcement learning algorithms?
    \item  How effective is each component in our proposed \emph{SAM-RL} algorithm?
    \item How effective is the \emph{model} update process leveraging the differentiable simulation and rendering?
\end{itemize}

\subsection{Experimental Setup}\label{sec:setup}

\subsubsection{Real World} We set up the real world experiment with two 7 DoF robotic arms as shown in Fig~\ref{fig:teaser}. One robot is the Franka Panda performing the manipulation task. The other robot is Flexiv Rizon~\cite{Flexiv} holding the RGB-D RealSense camera. We calibrate the camera intrinsic matrix and the hand-eye transformation between the camera and the Flexiv's end-effector. We also measure the relative transformation between two robots' bases in the world coordinate.

\begin{figure}[tbh!]
  \begin{center}
   \includegraphics[width=0.95\linewidth]{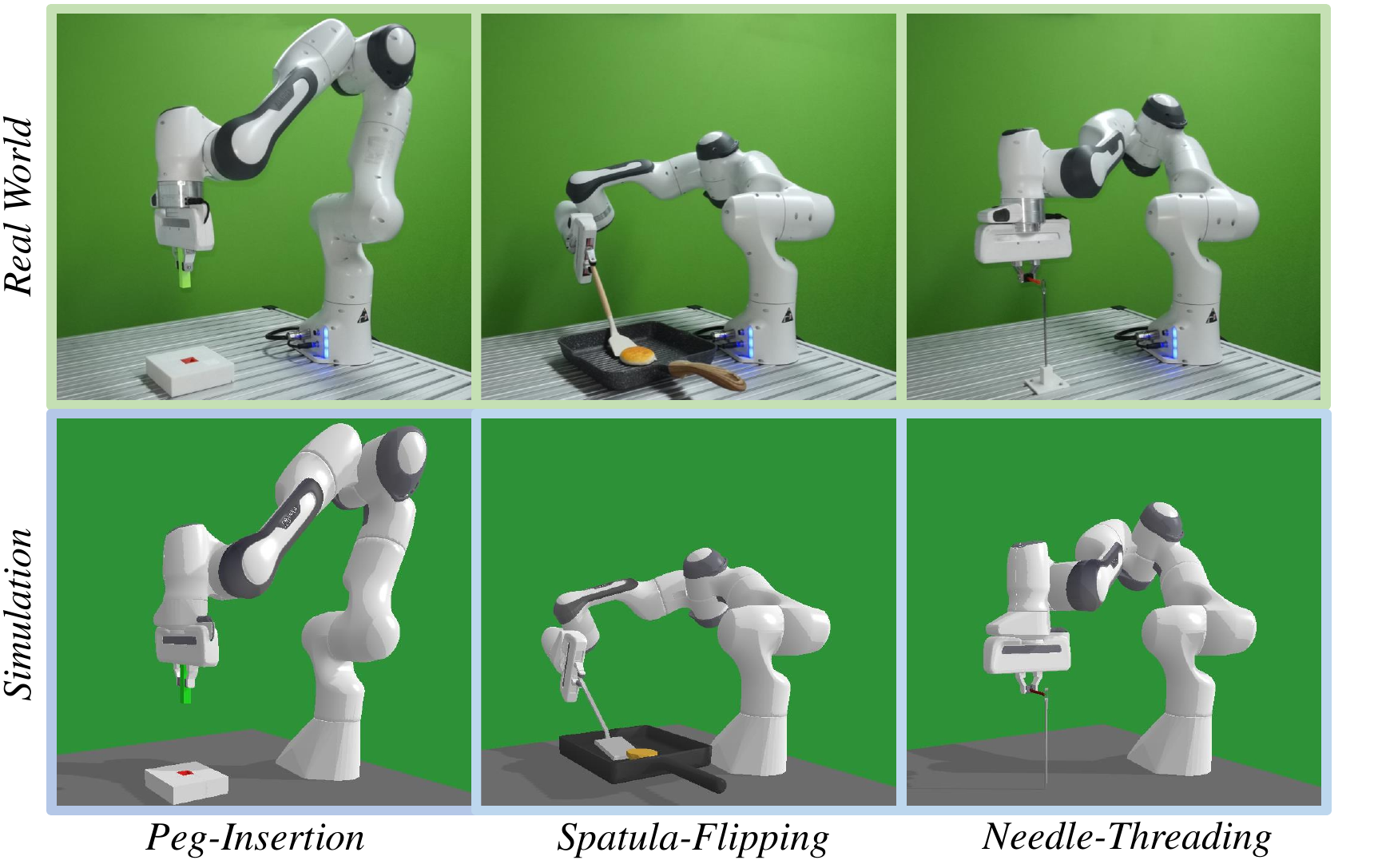}
  \end{center}
      \caption{Visualization of experiment setup for \emph{Peg-Insertion}, \emph{Spatula-Flipping}, and \emph{Needle-Threading}.}
\label{fig:robotsetting}
\end{figure}

\subsubsection{Simulation} In the simulation, we set the camera intrinsic matrix and relative transformations of the camera to Flexiv and Flexiv to Franka based on the real-world calibration results. We use PyBullet~\cite{coumans2021} to simulate the real world, which is different from our differentiable physics simulation for quantitative experiments. Objects are initialized with a random pose within manually defined bounds as described following.

\subsubsection{Differentiable Physics Simulation and Rendering} We combine the differentiable physics-based simulation and rendering to model the real world, update objects attributes, calculate expert trajectories, and optimize camera pose. We use NimblePhysics~\cite{werling2021fast} as differentiable simulation and Redner~\cite{Li:2018:DMC} as differentiable renderer.

\subsubsection{Manipulation Tasks}
\begin{figure}[h]
 \centering
 \includegraphics[width=0.7\linewidth ]{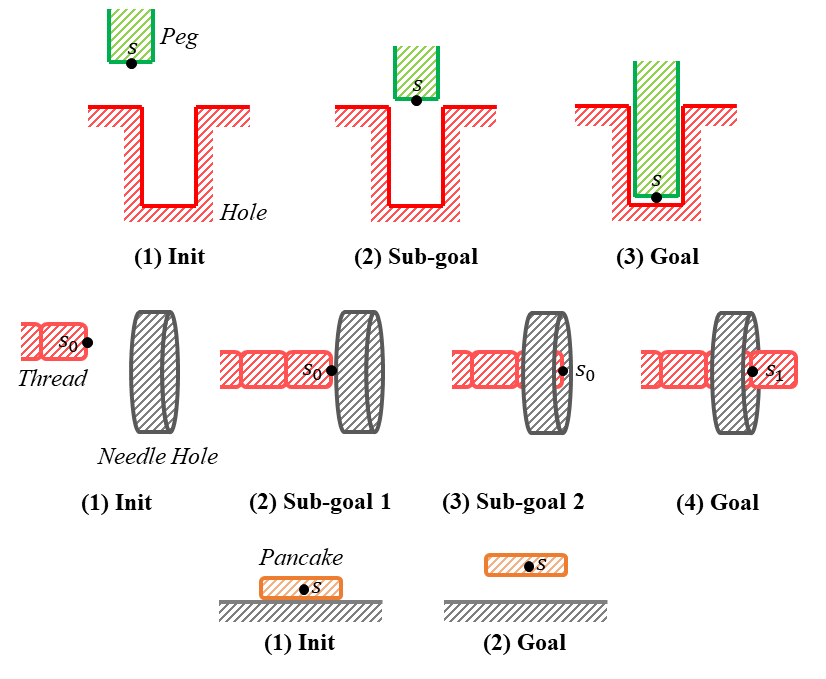}
  \caption{The goal of each task. The first row is \emph{Peg-Insertion}, the second is \emph{Needle-Threading}, and the third is \emph{Spatula-Flipping}.}
  \label{fig:sub_goal}
\end{figure}

We design three different manipulation tasks as shown in Fig.~\ref{fig:robotsetting}, and the goal of each task as shown in Fig.~\ref{fig:sub_goal}. 

\textbf{\emph{Peg-Insertion}}, which is inserting a peg into a hole. We assume the robot holds the peg tightly, so the state of the task is the translation of the peg, which is 3-dimension and could be calculated via robot forward kinematic. And the hole is initialized at a random location by $10 cm \times 10 cm$ both in simulation and real world. The task is considered successful if the peg is inserted into the hole. Our pipeline adopt an automatic termination function in real world by comparing the distance between the current state and goal state of the peg. The max number of action steps is 100.

\textbf{\emph{Spatula-Flipping}}, which is flipping a pancake with a spatula. The state of the task is the translation of the pancake, which is 3-dimension. The pancake is initialized at a random location by $2 cm\times 2 cm$ both in simulation and real world. The \emph{Spatula-Flipping} is successful if the pancake is lifted up by the spatula and then flipped. We track the position of the pancake via the method mentioned in Sec.~\ref{sec:updateM}. Our pipeline also adopt an automatic termination function for \emph{Spatula-Flipping} in real world by comparing the distance between the current state and goal state. The max number of action steps is 200.

\begin{figure}[h]
 \centering
 \includegraphics[width=0.4\linewidth ]{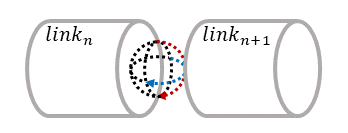}
  \caption{There are two revolute joints between links, rotate along the \emph{x} and \emph{y} axis (blue and red).} 
  \label{fig:thread}
\end{figure}

\textbf{\emph{Needle-Threading}}, which is threading a needle. In simulation, the thread is simulated using 10 links. There are two revolute joints (See Fig.~\ref{fig:thread}) between the links next to each other. The state of the task is the position of each link of the thread, which is 30-dimension in total. In the real world, the thread is manually randomly bent to initialize. While in simulation, for each revolute joint described in Fig.~\ref{fig:thread}, the state of the joint is initialized ranging from $-10^{\circ}$ to $10^{\circ}$. We manually decide success for \emph{Needle-Threading}, because it is hard to detect whether the thread is
through the needle hole automatically with high precision. The max number of action steps is 100.

\subsubsection{Training Details}
The size of the input images is $128 \times 128 \times 6$ (\emph{RGB-XYZ}), while the size of the action space is $3$, we only enable the translation of the Franka's end-effector and disable the rotation. For both the actor and critic ($Q$ function) of $\pi^{sim}$ and $\pi^{res}$, the network contains a CNN feature extractor and a MLP head. The feature extractor has 5 layers to extract a 128-dimensional vector from images, while the MLP head has 3 layers takes in the extracted vector, outputs the action or $Q$ value. Note that other inputs, like the predicted action for the critic network or the base action for residual actor, will be concatenated with the images and input to the feature extractor. 

In real world experiments, we train the residual policy network for 100 episodes for \emph{Needle-Threading} and 10 episodes for \emph{Peg-Insertion} and \emph{Spatula-Flipping}. To better reduce the sim-to-real gap, we also augment the training image $\mathcal{I}^{sim}$ by adding noise to the \emph{RGB} and \emph{XYZ} value of each pixel to imitate the sensor noise.

\subsection{Evaluating the Sensing-aware Function}
\subsubsection{Quantitative Results}
To evaluate the learned sensing-aware $Q$ function during Learn@Sim stage, we set up the following experiments inside the differentiable physics simulation. We use the same actor model~$\pi^{sim}$ trained with rendered images under multiple camera views, and evaluate the actor with and without leveraging the sensing-aware $Q$ function to optimize the camera pose during manipulation processes, denoted as~\emph{Ours} and~\emph{Ours(w/o pose-opt)}, respectively. We also train the same actor with rendered images under a fixed camera view, and evaluate the trained actor under the same camera view, denoted as~\emph{Fixed-View}. To make relatively fair compassion, we train the actor in~\emph{Fixed-View} to have the same prediction error in the training process as in~\emph{Ours}. We report the task success rate based on 100 experiments under these three settings. As shown in Tab.~\ref{tab:quantitative2}, our pipeline can find the informative view to benefit robot manipulation with the sensing-aware $Q$ function.
\begin{table}[tbh!]
\caption{Quantitative results of sensing-aware $Q$ function.}
\begin{center}
\begin{tabular}{c|ccc}
\toprule & \emph{Ours} & \emph{Ours(w/o pose-opt)} & \emph{Fixed View}\\
\midrule\midrule
\cellcolor{Gray}\emph{Peg Insertion}  &  \cellcolor{Gray}\bf0.82 & \cellcolor{Gray}0.64 &	\cellcolor{Gray}0.71 \\
\emph{Spatula Flipping} & \bf0.65 & 0.57 & 0.55\\
\cellcolor{Gray}\emph{Needle Threading}& \cellcolor{Gray}\bf0.70 & \cellcolor{Gray}0.46 &  \cellcolor{Gray}0.66 \\
\bottomrule
\end{tabular}
\end{center}
\label{tab:quantitative2}
\end{table}

\subsubsection{Qualitative Visualization}\label{sec:visQ}
\begin{figure}[h]
\centering
\subfigure[Visualization of the learned $Q$ function.]{
 \begin{minipage}[t]{0.85\linewidth}
    \includegraphics[width=\textwidth]{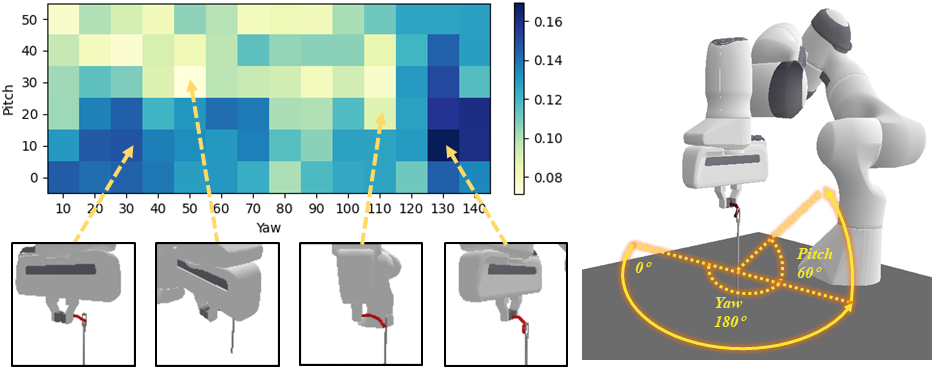}
    \centering
    \end{minipage}\label{fig:visQ}}
\subfigure[Visualization of the the prediction error of the learned imitation learning policy.]{
 \begin{minipage}[t]{0.85\linewidth}
    \centering
    \includegraphics[width=\textwidth]{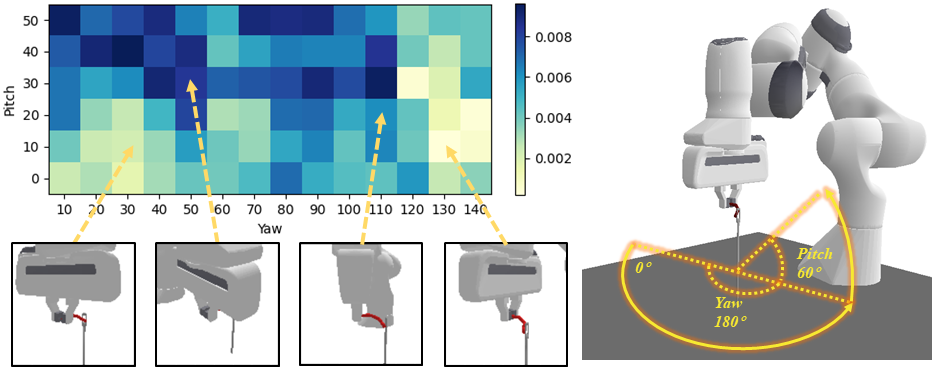}
   \end{minipage}\label{fig:visError}}
\caption{Qualitative visualization of sensing-aware $Q$ function for the \emph{Needle Threading}. Each pixel corresponds to a yaw and pitch value of the camera view.}
\end{figure}

We visualize the learned $Q$ function (Fig.~\ref{fig:visQ}) and the prediction error of the learned imitation learning policy (Fig.~\ref{fig:visError}) for the \emph{Needle-Threading} task. We gather rendered images $\mathcal{I}^{sim}(\mathcal{P}^{c,i})$ from multiple view points and send them to the actor~$\pi^{sim}$ to get associated actions $\pi^{sim}(\mathcal{I}^{sim}(\mathcal{P}^i))$, and get the $Q$ values $Q^{sim}(\mathcal{I}^{sim}(\mathcal{P}^{c,i}), \pi^{sim}(\mathcal{I}^{sim}(\mathcal{P}^{c,i})))$. It indicates that our $Q$ function learns a reasonable policy to select the camera viewpoint.

\subsubsection{Salient Map of Actor Network}
\begin{figure}[h]
\centering
\subfigure[\emph{Input Image}]{
 \begin{minipage}[t]{0.3\linewidth}
    \includegraphics[width=\textwidth]{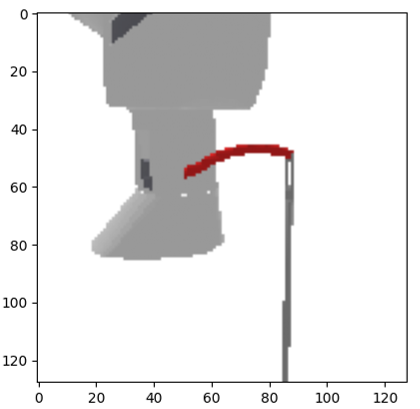}
    \centering
    \end{minipage}}
\subfigure[\emph{Salient Map}]{
 \begin{minipage}[t]{0.3\linewidth}
    \centering
    \includegraphics[width=\textwidth]{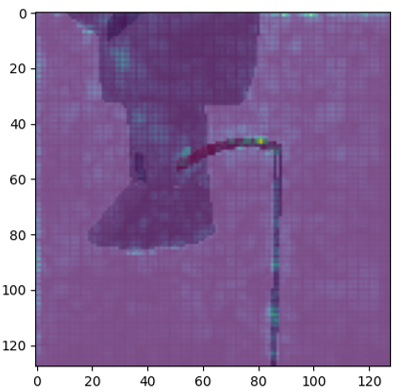}
   \end{minipage}}
\caption{The input image of the actor network and the corresponding salient map from the actor network}
\label{fig:salient}
\end{figure}

In order to have a better understanding about the actor network, we generate the salient map of the \emph{Needle-Threading} task as shown in Fig.~\ref{fig:salient} . Given an input image denoted as $\mathcal{I}^{sim}$, we calculate the prediction error denoted as
\begin{equation}
    \mathcal{L}_s=\|\hat{a}^{sim}-a^{sim}\|_2
\end{equation}
where $\hat{a}^{sim}$ and $a^{sim}$ are the predicted action and ground-truth action, respectively. We visualize the gradient $\partial \mathcal{L}_s / \partial \mathcal{I}^{sim}$ as the salient map. Fig.~\ref{fig:salient} indicates that the network pay more attentions to the thread's region, including the curved thread region and contact region between the thread and the needle.

\subsection{Comparing the SAM-RL with Model-free and Model-based Deep RL}
\begin{figure}[h]
\centering
\subfigure[Comparison with model-free and model-based RL in simulation on \emph{Peg-Insertion}]{
 \begin{minipage}[t]{0.45\linewidth}
    \includegraphics[width=\textwidth]{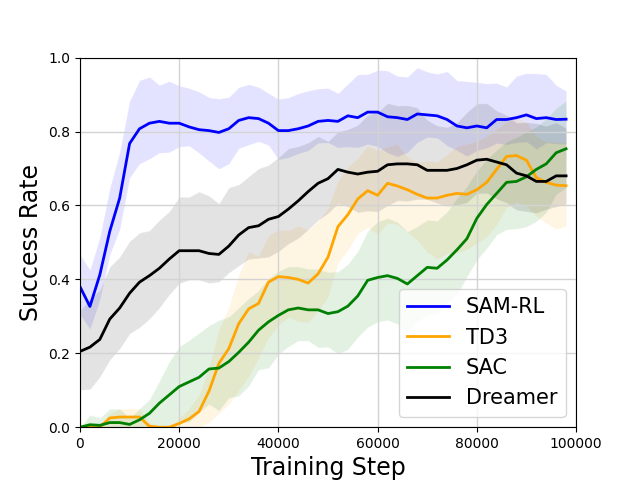}
    \centering
    \end{minipage}\label{fig:RLpeg}}
\subfigure[Comparison with model-free and model-based RL in simulation on \emph{Spatula-Flipping}]{
 \begin{minipage}[t]{0.45\linewidth}
    \includegraphics[width=\textwidth]{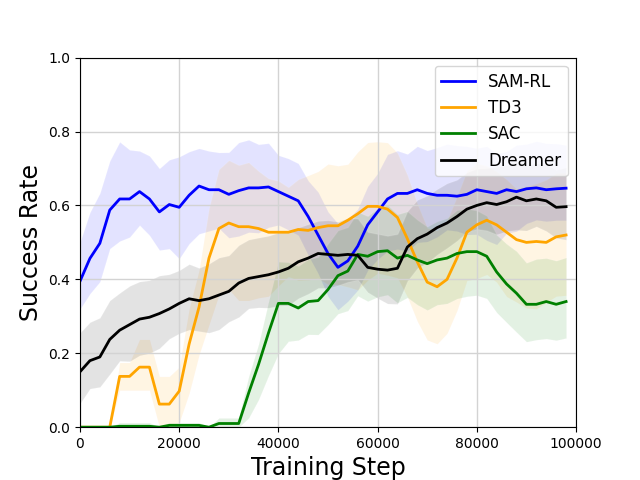}
    \centering
    \end{minipage}\label{fig:RLspatula}}
\subfigure[Comparison with model-free and model-based RL in simulation on \emph{Needle-Threading}]{
 \begin{minipage}[t]{0.45\linewidth}
    \includegraphics[width=\textwidth]{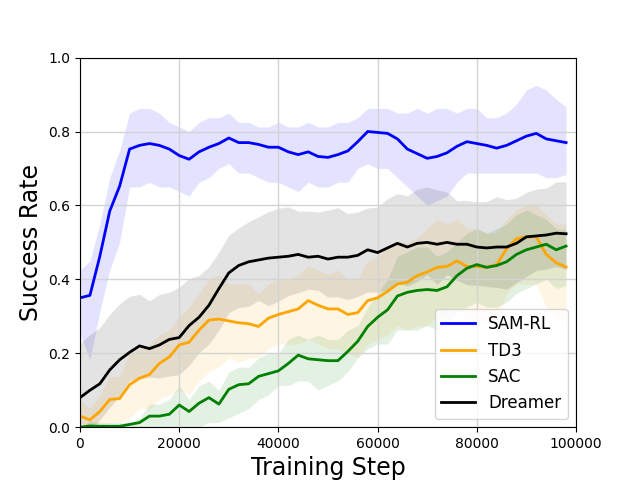}
    \centering
    \end{minipage}\label{fig:RL}}
\subfigure[Ablation studies in real world]{
 \begin{minipage}[t]{0.45\linewidth}
    \centering
    \includegraphics[width=\textwidth]{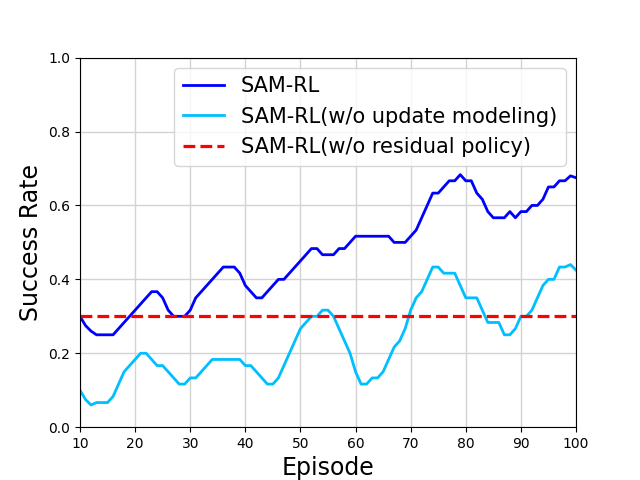}
   \end{minipage}\label{fig:ablation}}
\caption{Learning Curves of the \emph{Peg-Insertion}, \emph{Spatula-Flipping}, and \emph{Needle Threading}. The \emph{x-axis} shows the training steps/episodes and the \emph{y-axis} indicates the success rate.}
\end{figure}

We compare our \emph{SAM-RL} with common model-free deep RL algorithm \emph{TD3}~\cite{fujimoto2018addressing}, \emph{SAC}~\cite{haarnoja2018soft} and model-based deep RL algorithm \emph{Dreamer}~\cite{hafner2019dream}. In this experiment, we adopt pybullet as the ``real world". For \emph{TD3}, \emph{SAC}, and \emph{Dreamer}, we directly train the robot in pybullet. The observation including an \emph{RGB-XYZ} image and the position of the robot's end-effector. The action space is the translation of the robot's end-effector, which is 3 dimensions. The reward function of \emph{Peg-Insertion} is the distance between the peg and the hole. The reward function of \emph{Needle-Threading} is the distance between the thread and needle hole. The reward function of \emph{Spatula-Flipping} is the position of the pancake among \emph{z-axis} which is perpendicular to the ground. For \emph{SAM-RL}, we develop the \emph{model} via the differentiable physics-based simulation and rendering and use the \emph{model} to complete tasks in pybullet~(``real world"). Every time the environment is initialized or reset, the position and shape of the thread are randomly set within a certain range as described in Sec.~\ref{sec:setup}. Fig.~\ref{fig:RLpeg}, Fig.~\ref{fig:RLspatula}, and Fig.~\ref{fig:RL} show the average success rate of the \emph{SAM-RL}, \emph{SAC}, \emph{TD3}, and \emph{Dreamer} during the training stage. Take \emph{Needle-Threading} as an example, after training 100k steps, the average success rate of \emph{SAC} models and \emph{TD3} models are about 50\%, while \emph{Dreamer} is about 60\%. However, our pipeline achieves a success rate of around 80\% after 8k training steps. It indicates that our proposed \emph{SAM-RL} is significantly sample-efficient compared to model-free deep reinforcement learning approaches and existing model-based method.

\subsection{Ablation: Evaluating Components of SAM-RL}
We apply \emph{SAM-RL} in the real robot system and conduct two ablation study experiments.  
\begin{itemize}
    \item We remove the component of updating the \emph{model} $\mathcal{M}$ by comparing $\mathcal{I}^{real}$ and $\mathcal{I}^{sim}$ and denote the experiment as \emph{SAM-RL~(w/o update modeling)}.
    \item We remove the residual policy used to address the sim2real gap and directly execute the predicted action $a^{sim}$ in the real world denoted as \emph{SAM-RL~(w/o residual policy)}.
\end{itemize}
Note that \emph{SAM-RL~(w/o residual policy)} directly apply $\pi^{sim}$ in real world, have no need to train, so it has a constant success rate.
We post the result of \emph{Needle Threading} in Fig.~\ref{fig:ablation}. It shows that these components play important roles in our pipeline, and removing these components results in significant performance decreases.

\subsection{Evaluating the Model Updating via Differentiable Simulation and Rendering}

\subsubsection{Pose and Texture} With differentiable simulation and rendering, we can update the texture of the \emph{model} $\mathcal{M}$.
And also, we can update the pose of the \emph{model} $\mathcal{M}$ during the manipulation. For \emph{Peg-Insertion}, the \emph{model}'s pose is the translation of the hole on the table, which is 2-dimension (\emph{x} and \emph{y}). We only need to update this at the beginning of each episode because the location of the hole is randomly initialized and fixed during manipulation. For \emph{Spatula-Flipping}, The \emph{model}'s pose is 5-dimension, include the translation of the pancake, which is 3-dimension (\emph{x}, \emph{y}, and \emph{z}), and rotation of the pancake, which is 2-dimension (rotate among \emph{x-axis} and \emph{y-axis}) as the pancake is symmetry among the \emph{z-axis}. For \emph{Needle-Threading}, the thread is simulated using 10 links as described in Sec.~\ref{sec:setup}. There 
are two revolute joints (See Fig.~\ref{fig:thread}) between the links next to each other. So the \emph{model}'s pose is the states of the joints, which is 20-dimension in total. 

\begin{figure}[h]
 \centering
 \includegraphics[width=1.0\linewidth ]{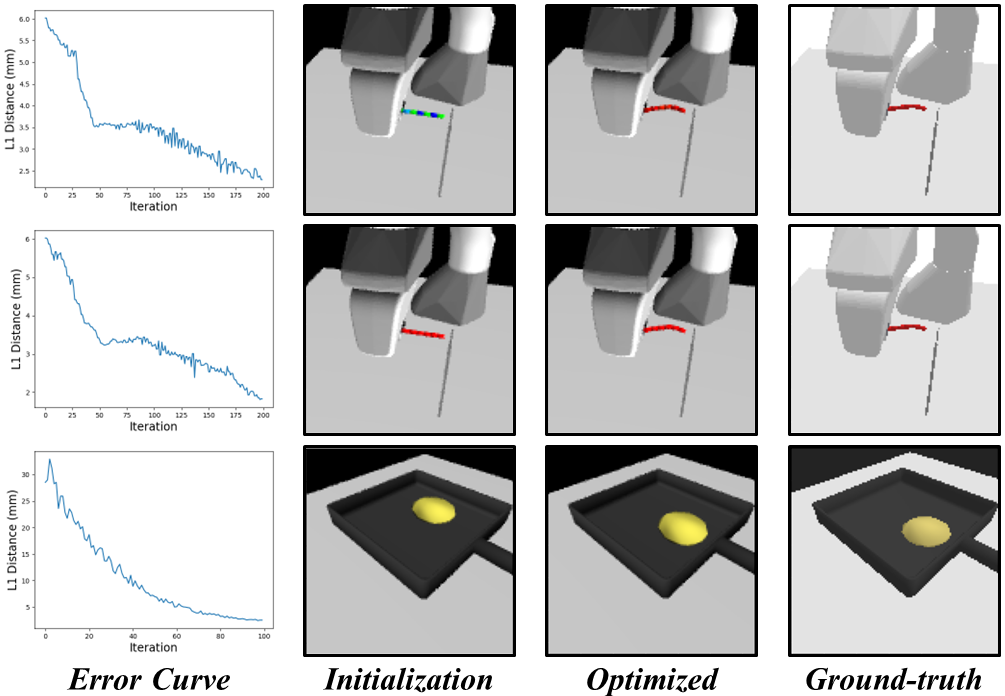}
  \caption{The initialization and ground-truth image of the \emph{model} updating for \emph{Needle-Threading} and \emph{Spatula-Flipping}. And the $L_1$ distance of the estimated and ground-truth state during \emph{model} updating.} 
  \label{fig:ba}
\end{figure}

As shown in Fig.~\ref{fig:ba}, we demonstrate the mean $L_1$ distance between the estimated and ground-truth position of the thread's links for \emph{Needle-Threading}, and the $L_1$ distance between the estimated and ground-truth position of the pancake. To get the ground-truth position of the thread and pancake, we use the image rendered by Pybullet rendering, which is different from the differential rendering, as the ground-truth image. It indicate that our pipeline can reduce the distance between the estimated and the ground-truth state and texture by \emph{model} updating via differentiable rendering. The first row shows the experiment where the thread~(in the left most image) has a different position and different color from the ground truth image~(as shown in the right most). The second row reflects the experiment where the thread has the same color but different position from the ground truth image. The last row shows the experiment where the pose of the pancake is different from the ground truth image.

Note that to better evaluate the performance of the \emph{model} updating, we make the distance between the initialization and ground-truth larger. So it takes 200 iterations for \emph{Needle-Threading} and 100 iterations for \emph{Spatula-Flipping} to optimize. In practice, one action step can only cause a little change to the thread and pancake, so we only take 20 iterations for the thread updating and 10 iterations for the pancake updating at each action step.

\subsubsection{Physical Attributes}
\begin{figure}[h]
 \centering
 \includegraphics[width=1.0\linewidth ]{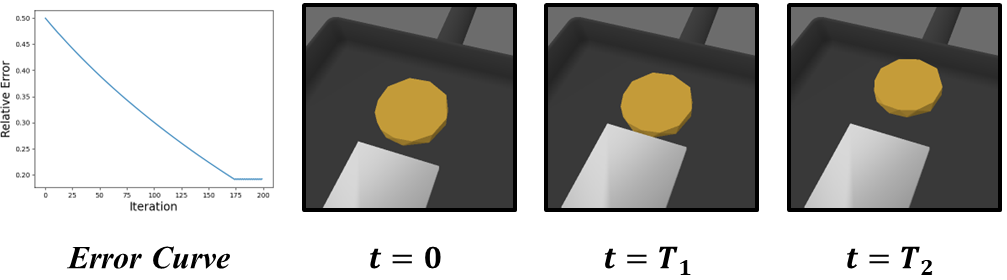}
  \caption{At timestep $t=0$, the robot start to push the pancake with the spatula until timestep $t=T_1$, then the robot stop moving and the pancake keep sliding until timestep $t=T_2$.} 
  \label{fig:phy}
\end{figure}

Next we describe how to update objects' physical attributes (mass and inertial) in detail. We still use Pybullet to simulate the real world. We first use the spatula to push the pancake for $T_1$ time steps, then we stop moving the spatula and the pancake keeps sliding until time step $T_2$~(See Fig.~\ref{fig:phy}). We use the robot to manipulate the spatula in torque control mode. So with the same robot action, the impulse provided by the robot is constant, and the pancake pose at time step $T_2$ is influenced by the mass value of the pancake. We update the object's physical attributes (mass and inertial) via the differentiable physics simulation. As shown in the learning curve, the relative error of the pancake's mass value can be reduced from $50\%$ to around $20\%$ after the optimization.

\subsubsection{Real World}
\begin{figure}[H]
 \centering
 \includegraphics[width=0.8\linewidth ]{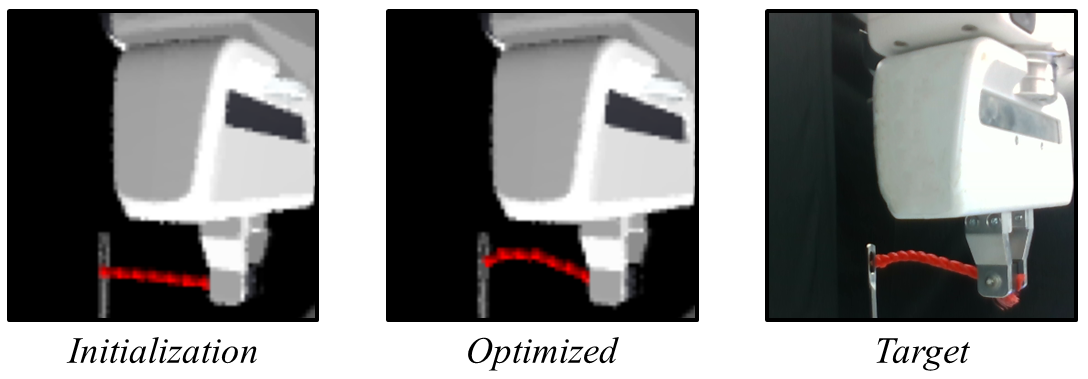}
  \caption{The rightmost image is the real image, the target image for \emph{model} update. The leftmost image and middle image show the initial state of the thread and the updated thread leveraging the differentiable rendering.} 
\label{fig:update}
\end{figure}

We also demonstrate the \emph{model} updating via differentiable rendering in the real world. (See Fig.~\ref{fig:update})

\section{Conclusion}
\label{sec:conclusion}
We propose a sensing-aware model-based reinforcement learning called \emph{SAM-RL} leveraging the
differentiable physics-based simulation and rendering. \emph{SAM-RL} automatically updates the \emph{model} by comparing the raw observations between simulation and the real world, and produces the policy efficiently. It also allows robots to select an informative viewpoint to better monitor the task process. We apply the system to robotic assembly, tool manipulation, and deformable object manipulation tasks. Extensive experiments in the simulation and the real world demonstrate the effectiveness of our proposed learning framework.

\section{Acknowledgement}
This work was in part supported by the National Key R\&D Program of China (2021ZD0110700), Shanghai Municipal Science and Technology Major Project (2021SHZDZX0102), Shanghai Qi Zhi Institute, Shanghai Science and Technology Commission (21511101200), and a startup grant from National University of Singapore.

{\small
\bibliographystyle{IEEEtranN}
\bibliography{references}
}

\end{document}